%% file: main.tex
\pdfoutput=1

\documentclass[11pt]{article}

\usepackage[]{ACL2023}

\usepackage{times}
\usepackage{latexsym}

\usepackage[T1]{fontenc}

\usepackage[utf8]{inputenc}

\usepackage{microtype}

\usepackage{inconsolata}

\usepackage{caption}
\usepackage{graphicx}
\usepackage{algpseudocode}
\usepackage{algorithm}
\usepackage{pdflscape}
\usepackage{booktabs}
\usepackage{comment}
\usepackage{enumitem}
\usepackage{xspace}
\usepackage{multirow,multicol}
\usepackage{tablefootnote}
\usepackage{threeparttable}
\usepackage{subcaption} 
\usepackage{tikz}
\usetikzlibrary{shapes.geometric, arrows}
\usepackage{float}

\newcommand{\dsname}{Naamapadam\xspace}
\newcommand{\modelname}{IndicNER\xspace}

\newcommand{\viz}{\textit{viz.}\xspace}

\newcommand{\cbullet}{\noindent $\bullet\;$}

\DeclareRobustCommand{\indicWords}[1]{
  \raisebox{-\dp\strutbox}{
    \includegraphics[page=#1]{indicWords}
  }
}

\title{\dsname: A Large-Scale Named Entity Annotated Data for Indic Languages}

\author{
        Arnav Mhaske$^{1,2}$\Thanks{\;Equal contribution}  \hspace{0.2cm} Harshit Kedia$^{1,2}$\footnotemark[1] \\   
           \textbf{Sumanth Doddapaneni}$^{1,2}$ \hspace{0.2cm} \textbf{Mitesh M. Khapra}$^{1,2}$ \hspace{0.2cm} \textbf{Pratyush Kumar}$^{1,2,3}$ \\
        \textbf{Rudra Murthy V}$^{2,4}$\Thanks{\;Project leads; correspondence to rmurthyv@in.ibm.com, ankunchu@microsoft.com} \hspace{0.2cm} \textbf{Anoop Kunchukuttan}$^{1,2,3}$\footnotemark[2]          
    \\ \\
    $^1$Indian Institute of Technology Madras \hspace{0.2cm} 
    $^2$AI4Bharat \\ 
    $^3$Microsoft India \hspace{0.2cm} 
    $^4$IBM Research India \hspace{0.2cm}
}

\begin{document}
\maketitle
\begin{abstract}
We present, \textit{{\dsname}}, the largest publicly available Named Entity Recognition (NER) dataset for the 11 major Indian languages from two language families. The dataset contains more than 400k sentences annotated with a total of at least 100k entities from three standard entity categories (Person, Location, and, Organization) for 9 out of the 11 languages. The training dataset has been automatically created from the Samanantar parallel corpus by projecting automatically tagged entities from an English sentence to the corresponding Indian language translation. We also create manually annotated testsets for 9 languages. We demonstrate the utility of the obtained dataset on the {\dsname}-test dataset.  We also release \textit{{\modelname}}, a multilingual IndicBERT model fine-tuned on {\dsname} training set. {\modelname} achieves an F1 score of more than $80$ for $7$ out of $9$ test languages. The dataset and models are available under open-source licences at \url{https://ai4bharat.iitm.ac.in/naamapadam}.
\end{abstract}

\input{sections/intro}

\input{sections/related}

\input{sections/mining}

\input{sections/human_labeled}

\input{sections/approach}

\input{sections/experiment}

\input{sections/results}

\input{sections/conclusion}

\section*{Limitations}


This work applies to languages that have a modest amount of data in parallel with English and are represented in pre-trained language models. These are typically high to mid-resource languages. Very low-resource languages might not have enough parallel corpora to extract sufficient NER training data. With limited parallel data and/or limited representation in pre-trained LMs, it will be difficult to get high-quality word alignments for projection. We use span-based annotation projection to alleviate word alignment errors to some extent.   

\section*{Ethics Statement}

The annotations are collected on a publicly available dataset and will be released publicly for future use. Some of these datasets originate from webcrawls and we do not make any explicit attempt to identify any biases in these datasets and use them as-is. All the datasets used have been cited. All the datasets created as part of this work will be released under a CC-0 license\footnote{\url{https://creativecommons.org/publicdomain/zero/1.0}} and all the code and models will be release under an MIT license.\footnote{\url{https://opensource.org/licenses/MIT}}

The annotations in the testset were mostly contributed by volunteers interested in contributing to building a benchmark NER dataset. The volunteers were not made any payment and worked  \textit{pro bono}. Some annotators were paid for their services.  These language experts were paid a competitive monthly salary to help with the task. The salary was determined based on the skill set and experience of the expert and adhered to the norms of the government of our country. The annotators were made aware that the annotations would be made publicly available. The annotations contains no personal information.

\bibliography{anthology,custom}
\bibliographystyle{acl_natbib}

\input{sections/appendix.tex}

\end{document}

%% file: sections/intro.tex
\section{Introduction}

\begin{table*}[!htb]
\centering
\small
\begingroup
\setlength{\tabcolsep}{4pt} 
\renewcommand{\arraystretch}{1} 
\begin{tabular}{lrrrrrrrrrrr}
\toprule 
 \multicolumn{1}{c}{} & \multicolumn{1}{c}{\textbf{as}} & \multicolumn{1}{c}{\textbf{bn}} & \multicolumn{1}{c}{\textbf{gu}} & \multicolumn{1}{c}{\textbf{hi}} & \multicolumn{1}{c}{\textbf{kn}} & \multicolumn{1}{c}{\textbf{ml}} & \multicolumn{1}{c}{\textbf{mr}} & \multicolumn{1}{c}{\textbf{or}} & \multicolumn{1}{c}{\textbf{pa}} & \multicolumn{1}{c}{\textbf{ta}} & \multicolumn{1}{c}{\textbf{te}} \\ 
 \midrule
\textbf{\dsname} & 5.0K & 1.6M & 769.3K & 2.2M & 658K & 1.0M & 735.0K & 190.0K & 880.2K & 745.2K & 751.1K \\ 
\textbf{WikiANN} & 218 & 12K & 264 & 7.3K & 220 & 13K & 7.3K & 265 & 211 & 19.7K & 2.4K \\ 
\textbf{FIRE-2014} & - & 6.1K & - & 3.5K & - & 4.2K & - & - & - & 3.2K & - \\
\textbf{CFILT} & - & - & - & 262.1K & - & - & 4.8K & - & - & - & - \\
\textbf{MultiCoNER} & - & 9.9K & - & 10.5K & - & - & - & - & - & - & - \\
\textbf{MahaNER} & - & - & - & - & - & - & 16K & - & - & - & - \\
\textbf{AsNER}$^\phi$ & $~$6K & - & - & - & - & - & - & - & - & - & - \\
\bottomrule
\end{tabular}%
\endgroup
\caption{Comparison of Indian language Named Entity training dataset statistics (total number of named entities), For all datasets, the statistics include only LOC, PER and ORG named entities. $\phi$ - While the dataset contains a total of 29K entities, most of the examples are gazetteer entries without sentence context.}
\label{num_ent_table}
\end{table*}

Named Entity Recognition (NER) is a fundamental task in natural language processing (NLP) and is an important component for many downstream tasks like information extraction, machine translation, entity linking, co-reference resolution, etc. The most common entities of interest are \texttt{person}, \texttt{location}, and, \texttt{organization} names, which are the focus of this work and most work in NLP. Given high-quality NER data, it is possible to train good-quality NER systems with existing technologies \cite{devlin-etal-2019-bert}. For many high-resource languages, publicly available annotated NER datasets \cite{tjong-kim-sang-2002-introduction,tsang2003introduction,pradhan-etal-2013-towards,benikova-etal-2014-nosta} as well as high-quality taggers \cite{wang-etal-2021-automated,li-etal-2020-dice} are available. 

\begin{figure}[ht!]
    \centering
    \includegraphics[width=\columnwidth]{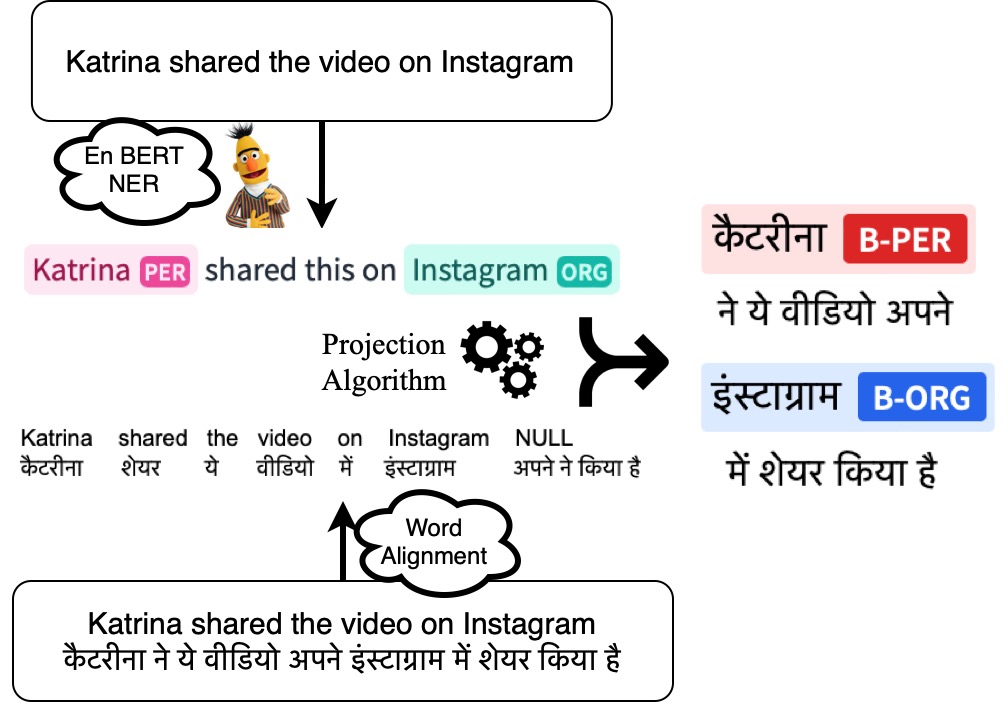}
    \caption{Illustration of Named Entity projection. We perform (i) NER with fine-tuned English BERT model, followed by (ii) word alignment between parallel sentence pair and (iii) projection of the English entities onto Indic sentence.}
    \label{fig:alignment}
\end{figure}

However, most Indic languages do not have sufficient labeled NER data to build good-quality NER models. All existing NER corpora for Indic languages have been manually curated \cite{FIRE2014,murthy-etal-2018-judicious,pathak-etal-2022-asner,https://doi.org/10.48550/arxiv.2204.13743,shervin-eta-al-2022-semeval,litake-EtAl:2022:WILDRE6}. Given the number of languages, the expenses and the logistical challenges, these datasets are limited on various dimensions \viz corpus size, language coverage, and broad domain representation. In recent years, zero-shot cross-lingual transfer from pre-trained models, fine-tuned on task-specific training data in English has been proposed as a way to support various language understanding tasks for low-resource languages \cite{pmlr-v119-hu20b}. However, this approach is more suitable for semantic tasks and the cross-lingual transfer does not work as well for syntactic tasks like NER when transferring across distant languages like English and Indian languages \cite{wu-dredze-2019-beto,K2020Cross-Lingual,ruder-etal-2021-xtreme}. Hence, there is a need for in-language NER training data for Indic languages.

In recent years, the paradigm of mining datasets from publicly available data sources has been successfully applied to various NLP tasks for Indic languages like machine translation \cite{ramesh-etal-2022-samanantar}, machine transliteration \cite{https://doi.org/10.48550/arxiv.2205.03018} and many natural language understanding and generation tasks \cite{kakwani2020indicnlpsuite,kumar-etal-2022-indicnlg}. These approaches have led to the creation of large-scale datasets and models with broad coverage of Indic languages in a short amount of time. Taking inspiration from these successes, we also explore the automatic creation of NER datasets by utilizing publicly available parallel corpora for Indian languages and high-quality English-named entity taggers. In this work, we undertake the task of building large-scale NER datasets and models for all major Indic languages.

The following are the contributions of our work: 

\cbullet We build \dsname\footnote{\dsname means named entity in Sanskrit}, the largest publicly available NER dataset for Indic languages for 11 languages from \textbf{2} language families. \dsname contains \textbf{5.7M} sentences and \textbf{9.4M} entities across these languages from three categories: \texttt{PERSON}, \texttt{NAME}, and \texttt{LOCATION}. This is significantly larger than other publicly available NER corpora for Indian languages in terms of the number of named entities and language coverage. Table \ref{num_ent_table} compares \dsname with other NER datasets. 

\cbullet We create the \dsname test set, containing human-annotated test sets for $9$ languages on general domain corpora, that can help in benchmarking NER models for Indic languages. Existing testsets are limited to fewer languages or are domain-specific.  

\cbullet We also train a multilingual NER model, IndicNER, supporting 11 Indic languages. Our models achieve more than 80\% F1 score on most languages on the \dsname test set. To the best of our knowledge, no publicly available NER models exist for Indian languages. 

\cbullet We create the NER training corpora by projecting annotations from English sentences to their Indic language translations in parallel corpora. We show that the projection approach is better than approaches based on zero-shot transfer or teacher-student training. This allows for the inexpensive creation of data, at scale, while maintaining high quality. Hence, we recommend the use of a projection approach compared to these approaches when a reasonable amount of parallel corpora is available. This is a valid assumption for many mid-resource languages which today lack good NER models. 


%% file: sections/related.tex
\section{Related Work}


We discuss the state of NER datasets for Indian languages and common methods used to improve NER for low-resource languages. 



 
\subsection{NER data for Indian languages}

Very limited NER corpora are available for Indian languages. They are mostly small in size and do not cover all major Indian languages. The FIRE-2014 dataset \cite{FIRE2014} is available for 4 languages. It was created by collecting sentences/documents from Wikipedia, blogs, and, online discussion forums.  The WikiAnn dataset \cite{pan2017cross} is available for around 16 Indian languages - but these are all Wikipedia article titles that are not representative of natural language sentences and the annotations are very noisy. Moreover, the examples are Wikipedia article titles and are not representative of natural language sentences. \citet{https://doi.org/10.48550/arxiv.2204.13743} contributed the largest human-annotated dataset for Hindi (CFILT-Hindi) in terms of volume and diversity with over 100k sentences, all annotated by a single expert individual over a span of several years. There are a few small datasets for Indian languages: CFILT-Marathi \cite{murthy-etal-2018-judicious}, MahaNER \cite{litake-EtAl:2022:WILDRE6}, AsNER \cite{pathak-etal-2022-asner} and MultiCoNER \cite{shervin-eta-al-2022-semeval}. In contrast, \dsname has greater language coverage and is much larger compared to other datasets. It is also representative of general domain text.   
 
\subsection{Annotation Projection} 
Named entity corpora can be created for low-resource languages by projecting named entity annotations from sentences in the source language (high-resource) to the corresponding words in the translated sentence in the target language (low-resource). \citet{yarowsky-etal-2001-inducing} first demonstrated how annotations can be projected using word alignments given parallel corpora between two languages. In addition to word alignments, projection can also be based on matching tokens via translation and entity dictionaries as well as transliteration \cite{zhang-etal-2016-bitext,jain-etal-2019-entity}. \citet{agerri2018building} extended this approach to multiple languages by utilizing multi-way parallel corpora to project named entity labels from multiple source languages to the target language. 
When parallel corpus is not available, but good quality MT systems are available, annotated corpora in one language can be translated to another language followed by annotation projection \cite{jain-etal-2019-entity,shah2010synergy}.  Bilingual dictionaries or bilingual embeddings have been used as a cheap alternative for translation in low-resource scenarios \cite{mayhew2017cheap,xie-etal-2018-neural}. The WikiAnn project creates `\textit{silver standard}' NER corpora using a weakly supervised approach leveraging knowledge bases and cross-lingual entity links to project English entity tags to other languages \cite{pan2017cross}. Given the availability of sufficient parallel corpora for major Indian languages \cite{ramesh-etal-2022-samanantar}, we use the annotation projection approach for building Indian language NER datasets.

\subsection{Zero-shot Cross-lingual Transfer}

This is a popular method for low-resource languages that relies on shared multilingual representations to help low-resource languages by transferring information from high-resource language NER models. Particularly, NER models finetuned on pre-trained language models like mBERT \cite{devlin-etal-2019-bert}, XLM-RoBERTa \cite{conneau-etal-2020-unsupervised} for high resource languages are used to tag low-resource language sentences (zero-shot NER). \citet{pires-etal-2019-multilingual} demonstrate that multilingual models perform well for zero-shot NER transfer on related languages. However, zero-shot performance is limited for distant languages \citet{wu-dredze-2019-beto}, particularly when there are structural/word order differences between the two languages \cite{K2020Cross-Lingual}. Unlike many other NLP tasks, zero-shot cross-lingual NER has seen only limited benefit from recent advances in cross-lingual representation learning \cite{ruder-etal-2021-xtreme}. To overcome this limitation, a knowledge distillation approach has been proposed to create synthetic in-language training data \cite{wu-etal-2020-single}. Here, the source language \textit{teacher} NER model is used to create distillation data in the target language via zero-shot cross-lingual transfer, which is used to train a target language model. We find that the projection-based approaches outperform zero-shot transfer and knowledge distillation approaches.







%% file: sections/mining.tex
\section{Mining NER Corpora}
\label{sec:mining}

Following \citet{yarowsky-ngai-2001-inducing,Yarowsky:2001:IMP:1073336.1073362}, our method for building NER corpora is based on projecting NER tags from the English side of an English-Indic language parallel corpora to the corresponding Indic language words. For our work, we use the Samanantar parallel corpus \cite{ramesh-etal-2022-samanantar} which is the largest publicly available parallel corpora between English and 11 Indic languages. Figure \ref{fig:alignment} illustrates our workflow for extracting named entity annotated Indic sentences from an English-Indic parallel sentence pair. It involves the following stages: (a) tagging the English sentence with a high-accuracy English NER model (Sec \ref{sec:label-english}), (b) aligning English and Indic language words in the parallel sentence pair (Sec \ref{sec:word-alignment}), (c) projecting NER tags from the English sentence to Indic words using the word alignments (Sec \ref{sec:projection}). These stages are further described in this section.

\subsection{Tagging the English side}
\label{sec:label-english}
We tag the named entities on the English side of the parallel corpus using a publicly available, high-quality off-the-shelf English NER tagger. We evaluated various English taggers on the CoNLL dataset (see Table \ref{tab:en-baselines} for comparison). Since the parallel corpora contain a significant number of Indian named entities, we also performed a manual analysis 
to understand the taggers' performance on these entities. Based on these comparisons, we used the \textit{BERT-base-NER}\footnote{https://huggingface.co/dslim/bert-base-NER} model for tagging the English portion of the Samanantar parallel corpus. We ignore the \textit{MISC} tags predicted by the BERT-base-NER and focus on \texttt{PERSON}, \texttt{LOCATION}, and \texttt{ORGANIZATION} tags only. \texttt{MISC} is a very open-ended category and we found that it was not easy to reliably align \texttt{MISC} tagged entities from English to Indian languages.

\subsection{Word Alignment}
\label{sec:word-alignment}
For every sentence pair in the parallel corpus, we align English words to the corresponding Indic language words. We explore two approaches for learning word alignments: (a) GIZA++ \cite{och-ney-2003-systematic} with default settings, (b) 
Awesome-align \cite{dou-neubig-2021-word} finetuned on parallel corpora with  Translation Language Modeling and Self-training objectives. We use \textit{softmax} to normalize the alignment scores in our experiments.




\begin{figure*}[!htb]
    \centering
    \begin{minipage}[h]{0.45\textwidth}
        \centering
        \resizebox{0.95\textwidth}{!}{%
        \tikzset{every picture/.style={line width=0.75pt}} 
        \begin{tikzpicture}[x=0.75pt,y=0.75pt,yscale=-1,xscale=1]
        
            \draw (40,25) rectangle node[midway] {\scriptsize \textbf{LOCATION}} (120,48);
            \draw (80,60) node  [align=left] {Jharkhand};
            \draw (80,72) node  [align=left] {{\scriptsize 0}};
            
            \draw (140,60) node  [align=left] {chief};
            \draw (140,72) node  [align=left] {{\scriptsize 1}};
            
            \draw (200,60) node  [align=left] {minister};
            \draw (200,72) node  [align=left] {{\scriptsize 2}};
            
            \draw (230,25) rectangle node[midway] {\scriptsize \textbf{PERSON}} (330,48);
            \draw (260,60) node  [align=left] {Hemant};
            \draw (260,72) node  [align=left] {{\scriptsize 3}};
            
            \draw (320,60) node  [align=left] { [Soren] };
            \draw (320,72) node  [align=left] {{\scriptsize 4}};
            
            \draw (70,115) node  [align=left] {{\scriptsize 0}};
            \draw (70,130) node  [align=left] { \indicWords{113} };
            \draw (70,145) node  [align=left] {{\scriptsize (jharkhand)}};
            \draw (40,155) rectangle node[midway] {\scriptsize \textbf{LOCATION}} (100,175);
            
            \draw (110,115) node  [align=left] {{\scriptsize 1}};
            \draw (110,130) node  [align=left] { \indicWords{114} };
            \draw (110,145) node  [align=left] {{\scriptsize (ke)}};
            
            \draw (155,115) node  [align=left] {{\scriptsize 2}};
            \draw (155,130) node  [align=left] { \indicWords{115} };
            \draw (155,145) node  [align=left] {{\scriptsize (mukhyamaMtrI)}};
            
            \draw (210,115) node  [align=left] {{\scriptsize 3}};
            \draw (210,130) node  [align=left] { \indicWords{116} };
            \draw (210,145) node  [align=left] {{\scriptsize (hemant)}};
            
            \draw (250,115) node  [align=left] {{\scriptsize 4}};
            \draw (250,130) node  [align=left] { \indicWords{117} };
            \draw (250,145) node  [align=left] {{\scriptsize (soren)}};
            
            \draw (290,115) node  [align=left] {{\scriptsize 5}};
            \draw (290,130) node  [align=left] { \indicWords{118} };
            \draw (290,145) node  [align=left] {{\scriptsize (photo:)}};
            
            \draw (330,115) node  [align=left] {{\scriptsize 6}};
            \draw (330,130) node  [align=left] { \indicWords{119} };
            \draw (330,145) node  [align=left] {{\scriptsize (PTI)}};
            \draw (190,155) rectangle node[midway] {\scriptsize \textbf{PERSON}} (350,175);
            
            
            \draw[->,cyan ] (80, 80) -- (75,110);
            \draw[->,cyan ] (140, 80) -- (150,110);
            \draw[->,cyan ] (200, 80) -- (162,110);
            \draw[->,cyan ] (260, 80) -- (217,110);
            \draw[->,cyan ] (320, 80) -- (260,110);
            \draw[->,cyan ] (320, 80) -- (300,110);
            \draw[->,cyan ] (320, 80) -- (340,110);
            
            \draw[-> ] (70,110) -- (75, 80);
            \draw[-> ] (155,110) -- (145, 80);
            \draw[-> ] (155,110) -- (190, 80);
            \draw[-> ] (208,110) -- (250, 80);
            \draw[-> ] (250,110) -- (310, 80);
            
        \end{tikzpicture}%
        }
        \caption{Error in Word Alignments could lead to incorrect entity projection}
        \label{fig:projection}
    \end{minipage}
    \hfill
    \begin{minipage}[h]{0.45\textwidth}
        \centering
        \resizebox{0.95\textwidth}{!}{%
        \tikzset{every picture/.style={line width=0.75pt}} 
        
        \begin{tikzpicture}[x=0.75pt,y=0.75pt,yscale=-1,xscale=1]
        
            \draw (40,25) rectangle node[midway] {\scriptsize \textbf{LOCATION}} (120,48);
            \draw (80,60) node  [align=left] {Jharkhand};
            \draw (80,72) node  [align=left] {{\scriptsize 0}};
            
            \draw (140,60) node  [align=left] {chief};
            \draw (140,72) node  [align=left] {{\scriptsize 1}};
            
            \draw (200,60) node  [align=left] {minister};
            \draw (200,72) node  [align=left] {{\scriptsize 2}};
            
            \draw (230,25) rectangle node[midway] {\scriptsize \textbf{PERSON}} (330,48);
            \draw (260,60) node  [align=left] {Hemant};
            \draw (260,72) node  [align=left] {{\scriptsize 3}};
            
            \draw (320,60) node  [align=left] { [Soren] };
            \draw (320,72) node  [align=left] {{\scriptsize 4}};
            
            \draw (70,115) node  [align=left] {{\scriptsize 0}};
            \draw (70,130) node  [align=left] { \indicWords{113} };
            \draw (70,145) node  [align=left] {{\scriptsize (jharkhand)}};
            \draw (40,155) rectangle node[midway] {\scriptsize \textbf{LOCATION}} (100,175);
            
            \draw (110,115) node  [align=left] {{\scriptsize 1}};
            \draw (110,130) node  [align=left] { \indicWords{114} };
            \draw (110,145) node  [align=left] {{\scriptsize (ke)}};
            
            \draw (155,115) node  [align=left] {{\scriptsize 2}};
            \draw (155,130) node  [align=left] { \indicWords{115} };
            \draw (155,145) node  [align=left] {{\scriptsize (mukhyamaMtrI)}};
            
            \draw (210,115) node  [align=left] {{\scriptsize 3}};
            \draw (210,130) node  [align=left] { \indicWords{116} };
            \draw (210,145) node  [align=left] {{\scriptsize (hemant)}};
            
            \draw (250,115) node  [align=left] {{\scriptsize 4}};
            \draw (250,130) node  [align=left] { \indicWords{117} };
            \draw (250,145) node  [align=left] {{\scriptsize (soren)}};
            \draw (190,155) rectangle node[midway] {\scriptsize \textbf{PERSON}} (270,175);
            
            \draw (290,115) node  [align=left] {{\scriptsize 5}};
            \draw (290,130) node  [align=left] { \indicWords{118} };
            \draw (290,145) node  [align=left] {{\scriptsize (photo:)}};
            
            \draw (330,115) node  [align=left] {{\scriptsize 6}};
            \draw (330,130) node  [align=left] { \indicWords{119} };
            \draw (330,145) node  [align=left] {{\scriptsize (PTI)}};
            
            
            \draw[->,cyan ] (80, 80) -- (75,110);
            \draw[->,cyan ] (140, 80) -- (150,110);
            \draw[->,cyan ] (200, 80) -- (162,110);
            \draw[->,cyan ] (260, 80) -- (217,110);
            \draw[->,cyan ] (320, 80) -- (260,110);
            \draw[->,cyan ] (320, 80) -- (300,110);
            \draw[->,cyan ] (320, 80) -- (340,110);
            
            \draw[-> ] (75, 80) -- (70,110);
            \draw[-> ] (145, 80) -- (155,110);
            \draw[-> ] (190, 80) -- (155,110);
            \draw[-> ] (250, 80) -- (208,110);
            \draw[-> ] (310, 80) -- (250,110);
            
        \end{tikzpicture}%
        }
        \caption{Understanding how the intersection helps reduce errors}
        \label{fig:projection_intersection}
    \end{minipage}
\end{figure*}

\subsection{Projecting Named Entities}
\label{sec:projection}
The next step is the projection of named entity labels from English to the Indic language side of the parallel corpus using English-Indic language word alignment information. We want the entity projection algorithm to ensure the following: (1) adjacent entities of the same type should not be merged into one single entity, and (2) small errors in word alignment should not cause drastic changes in the final NER projection. To ensure these, we project the entities as a whole (\textit{i.e.,} the entire English entity phrase and not word by word) by identifying the minimal span of Indic words that encompass all the aligned Indic words. Word alignment errors could lead to incorrect named entity projection as shown in Figure \ref{fig:projection}. In this case, alignments in one direction are erroneous leading to wrong projection. We rely on the intersection of alignments in both directions to reduce alignment errors and thus ensure improved projection as illustrated in Figure \ref{fig:projection_intersection}. We show some examples of projections from Awesome-align in Appendix \ref{sec:alignment_examples}.

In Figure \ref{fig:projection} and \ref{fig:projection_intersection}, we use black arrows to indicate the alignment from Hindi to English direction and blue arrows to indicate the alignment from English to Hindi. The alignment from Hindi to English is correct. On the contrary, the alignment in English to Hindi direction suffers due to the presence of additional Hindi words. The word ‘Soren’ gets aligned to additional Hindi words ’photo’ and ‘PTI’ which are not part of \texttt{PERSON} named entity (Figure \ref{fig:projection}). In order to minimize such errors, we take advantage of bidirectional alignments. We take the intersection of alignments in both directions, which improves the precision of alignments and hence improves projection accuracy (Figure \ref{fig:projection_intersection}). We will include the description in the revised version. Figure \ref{sec:alignment_examples} is described in detail in Appendix C.

\begin{table*}[!htb]
\centering
\small
\begin{tabular}{lrrrrrrrr|r}
\toprule
\textbf{Language} & \multicolumn{1}{c}{\textbf{bn}} & \multicolumn{1}{c}{\textbf{gu}} & \multicolumn{1}{c}{\textbf{hi}} & \multicolumn{1}{c}{\textbf{kn}} & \multicolumn{1}{c}{\textbf{ml}} & \multicolumn{1}{c}{\textbf{mr}} & \multicolumn{1}{c}{\textbf{ta}} & \multicolumn{1}{c}{\textbf{te}}  & \multicolumn{1}{|c}{\textbf{Average}} \\ \midrule
\begin{tabular}[c]{@{}l@{}}\textbf{Awesome-align} \end{tabular} & 82.11 & 69.77 & \textbf{90.32} & 70.22 & \textbf{69.83} & \textbf{76.51} & \textbf{70.09} & 77.70 & \textbf{75.82} \\ 
\textbf{GIZA++} & \textbf{86.55} & \textbf{72.91} & 85.22 & \textbf{71.56} & 64.38 & 76.21 & 56.82 & \textbf{79.09} & 74.09 \\ 
\bottomrule
\end{tabular}
\caption{F1 scores from Different Projection Methods}
\label{table:projectionscores}
\end{table*}

\subsection{Sentence Filtering}

After NER projection, we apply the following filters to the tagged Indic sentences. 

\noindent \textbf{Sentences without Named Entities.} Many English sentences in the Samanantar corpus are not annotated with any entities. We retain only a small fraction of such sentences $(\approx1\%)$ for training the NER model so the model is exposed to sentences without any NER tags as well.

\noindent \textbf{Sentences with low-quality alignments.} We observe that most of the errors in the Indic-tagged corpus arise from word alignment errors. Hence, we compute a word alignment quality score for each sentence pair. This score is the product of the probability of each aligned word pair (as provided by the forward alignment model in the case of GIZA++ and the alignment model by awesome align) normalized by the number of words in the sentence. We retain the top 30-40\% sentences to create the final NER-tagged data for Indic languages (See Table \ref{filt_table} for filtered data statistics).   


\subsection{Qualitative Analysis}
\label{sec:alignmentQualitativeAnalysis}
To quantify the quality of the labeled data obtained, we select a small sample of $50$ sentences\footnote{this set is later expanded by annotating more sentences to form the test set} and obtain manual annotation for the $9$ languages namely \textit{Bengali, Gujarati, Hindi, Kannada, Malayalam, Marathi, Punjabi, 
 Tamil,} and, \textit{Telugu}. We also project the named entities on this small set of $50$ sentences using the projection approach discussed earlier. Since the ground truths are known, the F1 scores can be calculated. Table \ref{table:projectionscores} presents the F1 scores on the manually labeled set using various projection approaches. We observe that both GIZA++ and \textit{Awesome-align} word alignment approaches obtain similar performance. On average, \textit{Awesome-align} provides the best F1 scores, hence, moving forward, we consider the datasets from the \textit{Awesome-align} approach unless specified otherwise. 

\begin{table*}[!htb]
\centering
\small
\begin{tabular}{@{}lrrrrrrrrrrrr@{}}
\toprule
\multirow{2}{*}{\textbf{Lang.}} & \multicolumn{3}{c}{\textbf{Sentence Count}} & \multicolumn{3}{c}{\textbf{Train}} & \multicolumn{3}{c}{\textbf{Dev}} & \multicolumn{3}{c}{\textbf{Test}} \\ 
\cmidrule(lr){2-4}\cmidrule(lr){5-7}\cmidrule(lr){8-10}\cmidrule(lr){11-13}
 & \multicolumn{1}{c}{\textbf{Train}} & \multicolumn{1}{c}{\textbf{Dev}} & \multicolumn{1}{c}{\textbf{Test}} & \multicolumn{1}{c}{\textbf{Org}} & \multicolumn{1}{c}{\textbf{Loc}} & \multicolumn{1}{c}{\textbf{Per}} & \multicolumn{1}{c}{\textbf{Org}} & \multicolumn{1}{c}{\textbf{Loc}} & \multicolumn{1}{c}{\textbf{Per}} & \multicolumn{1}{c}{\textbf{Org}} & \multicolumn{1}{c}{\textbf{Loc}} & \multicolumn{1}{c}{\textbf{Per}} \\ 
 \midrule
bn & 961.7K & 4.9K & 607 & 340.7K & 560.9K & 725.2K & 1.7K & 2.8K & 3.7K & 207 & 331 & 457 \\
gu & 472.8K & 2.4K & 1.1K & 205.7K & 238.1K & 321.7K & 1.1K & 1.2K & 1.6K & 419 & 645 & 673 \\
hi & 985.8K & 13.5K & 867 & 686.4K & 731.2K & 767.0K & 9.7K & 10.2K & 10.5K & 521 & 613 & 788 \\
kn & 471.8K & 2.4K & 1.0K & 167.5K & 177.0K & 310.5K & 882 & 919 & 1.6K & 291 & 397 & 614 \\
ml & 716.7K & 3.6K & 974 & 234.5K & 308.2K & 501.2K & 1.2K & 1.6K & 2.6K & 309 & 482 & 714 \\
mr & 455.2K & 2.3K & 1.1K & 164.9K & 224.0K & 342.3K & 868 & 1.2K & 1.8K & 391 & 569 & 696 \\
pa & 463.5K & 2.3K & 993 & 235.0K & 289.8K & 351.1K & 1.1K & 1.5K & 1.7K & 408 & 496 & 553 \\
ta & 497.9K & 2.8K & 758 & 177.7K & 281.2K & 282.2K & 1.0K & 1.5K & 1.6K & 300 & 388 & 481 \\
te & 507.7K & 2.7K & 847 & 194.1K & 205.9K & 347.8K & 1.0K & 1.0K & 2.0K & 263 & 482 & 608 \\
\midrule
as & 10.3K & 52 & 51 & 2.0K & 1.8K & 1.2K & 18 & 5 & 3 & 11 & 7 & 6 \\
or & 196.8K & 993 & 994 & 45.6K & 59.4K & 84.6K & 225 & 268 & 386 & 229 & 266 & 431 \\
\bottomrule
\end{tabular}%
\caption{Statistics for the \dsname dataset. The testsets for  \texttt{as} and \texttt{or} are silver standard. Work on the creation of larger, manually annotated testsets is in progress for these languages.}
\label{CorpusStats}
\end{table*}

\subsection{Dataset Statistics}

Table \ref{CorpusStats} shows the statistics of the final \dsname dataset. We create train, dev, and, test splits. Testsets are then manually annotated as described later in Section \ref{sec:testset}. Most languages have training datasets of more than $100K$ sentences and $500K$ entities each. Some languages like Hindi have more than $1M$ sentences in the training set. Compared to other datasets (See Table \ref{num_ent_table}), the \dsname has a significantly higher number of entities. Even though the dataset is slightly noisy due to alignment errors, we hope that the large dataset size can compensate for the noise as has been seen in many NLP tasks \cite{bansal2022data}. 

We have manually annotated testsets of around 500-1000 sentences for most languages. The  \texttt{Assamese} and \texttt{Oriya} testsets are silver-standard (the named entity projections have not been verified yet). Work on the creation of larger, manually annotated testsets for these languages is in progress.

%% file: sections/human_labeled.tex
\section{Testset Creation}
\label{sec:testset}

We have created \dsname-test: manually annotated test set for Indian language NER evaluation. The \dsname-test comprises 500-1000 annotated sentences per language for 9 languages namely Bengali, Gujarati, Hindi, Kannada, Malayalam, Marathi, Punjabi, Tamil, and, Telugu. The annotators were provided sentences with named entity annotations obtained using the methodology described in Section \ref{sec:mining}. The annotators had to verify if the projected NER annotations were correct and rectify the annotations if incorrect. They were asked to follow the CoNLL 2003 annotation guidelines \cite{tjong-kim-sang-de-meulder-2003-introduction}. The human annotations were contributed by volunteers who are native language speakers. 

\subsection{Inter-Annotator Agreement}

\begin{table}[!htb]
\centering
\small
\begin{tabular}{@{}lrrr@{}}
\toprule
\multicolumn{1}{c}{\multirow{2}{*}{\textbf{Language}}} & \multicolumn{1}{c}{\multirow{2}{*}{\textbf{F1-Score}}} & \multicolumn{2}{c}{\textbf{Token-level Cohen's Kappa}} \\
\cmidrule{3-4}
\multicolumn{1}{c}{} & \multicolumn{1}{c}{} & \multicolumn{1}{c}{\textbf{All Tokens}} & \multicolumn{1}{c}{\textbf{Entity Tokens}} \\
\midrule
Bengali & 78.51 & 0.8506 & 0.7033 \\
Gujarati & 74.45 & 0.7965 & 0.6169 \\
Hindi & 93.60 & 0.9536 & 0.8996 \\
Kannada & 89.20 & 0.9217 & 0.8452 \\
Malayalam & 84.28 & 0.9006 & 0.8156 \\
Marathi & 88.20 & 0.9037 & 0.8047 \\
Punjabi & 60.01 & 0.6948 & 0.4605 \\
Tamil & 64.29 & 0.7176 & 0.5209 \\
Telugu & 78.40 & 0.8888 & 0.7850 \\
\bottomrule
\end{tabular}
\caption{IAA scores for 9 languages in the \dsname testset, We report F1-Score and token-level Cohen's Kappa for IAA.  \cite{interAnnotatorCite}. Token-level Cohen's Kappa* refers to the configuration where we consider tokens that are part of at least one mention according to at least one annotator \cite{ringland-etal-2019-nne}.}
\label{IAA}
\end{table}


We compute the inter-annotator agreement on a sample of two annotators for each language using Cohen's kappa coefficient \cite{cohen1960coefficient}. The scores are shown in  Table \ref{IAA}. They are all above 69\% signifying good-quality annotations. 


%% file: sections/approach.tex

%% file: sections/experiment.tex
\section{Experimental Setup}
We analyze the performance of models trained on the {\dsname-train} dataset with alternative approaches for low-resource NER and to models trained on publicly available datasets. To this end, we investigate the following research questions:

\begin{description}
    \item [RQ1:] Are models trained on data obtained from projection approach (\dsname-train) better than zero-shot and teacher-student models?
    \item [RQ2:] How does the model trained on publicly-available dataset fare against the model trained on {\dsname-train} data? We evaluate it on the {\dsname-test} set.
\end{description}

\subsection{Train Dataset}
In order to demonstrate the usefulness of our {\dsname-train} dataset, we fine-tune the mBERT model \cite{devlin-etal-2019-bert} on the {\dsname-train} data and test on {\dsname-test} set. We additionally fine-tune the mBERT model on the train split of publicly available datasets and test on {\dsname-test} set. We consider the following datasets in our experiments

\begin{itemize}
 \item \textbf{WikiANN}: We use the train split of the data released by \citet{rahimi-etal-2019-massively}. Due to the languages covered, this is the most widely used dataset. However, we observe the tagged data to be highly erroneous and does not contain complete sentences, but just titles. Appendix \ref{sec:wikiann} discusses the issues with the WikiNER dataset. 
 \item \textbf{FIRE-2014}: The FIRE-2014 dataset \cite{FIRE2014} contains named entity annotated dataset for Hindi, Bengali, Malayalam, Tamil, and, English languages. We train language-specific models on the train splits of these datasets and evaluate the performance on our test set. 
 \item \textbf{MultiCoNER}: We use the Hindi and Bengali named entity annotated data from \citet{shervin-eta-al-2022-semeval}.\footnote{\url{https://multiconer.github.io/}} 
 \item \textbf{CFILT}: We use the CFILT-HiNER dataset created for Named Entity Recognition in Hindi language \cite{https://doi.org/10.48550/arxiv.2204.13743}. The dataset was from various government information web pages, and newspaper articles. The sentences were manually annotated. We also use the CFILT-Marathi dataset created for Named Entity Recognition in Marathi \cite{murthy-etal-2018-judicious}.
 \item \textbf{MahaNER}: We use the Marathi named entity annotated data from \citet{litake-EtAl:2022:WILDRE6}.
\end{itemize}

For a fair comparison with models trained on our dataset, we include only \texttt{PERSON}, \texttt{LOCATION}, and, \texttt{ORGANIZATION} entities. The rest of the named entities if present (FIRE 2014, CFILT Marathi, MultiCoNER) are considered non-named entities.

\subsection{NER Fine-tuning}
Recently, sequence labeling via fine-tuning of pre-trained language models has become the norm \cite{devlin-etal-2019-bert,conneau-etal-2020-unsupervised,kakwani2020indicnlpsuite}. We fine-tune the pre-trained mBERT model \cite{devlin-etal-2019-bert} and report the results in our experiments. The input to the model is a sequence of sub-word tokens that pass through the Transformer encoder layers. The output from the transformer is an encoder representation for each token in the sequence. We take the encoder representation of the first sub-word (in case the word gets split into multiple sub-words) and is passed through the output layer. The output layer is a linear layer followed by the softmax function. The model is trained using cross-entropy loss. We use the \citet{dhamecha-etal-2021-role} toolkit for fine-tuning our models.  

\subsection{Baseline Comparison}

Our proposed approach can be seen as a cross-lingual approach since the training data is created by projection from English to Indic sentences. Hence, we compare the performance of our model with zero-shot learning \cite{pires-etal-2019-multilingual} and teacher-student learning \cite{wu-etal-2020-single}. We describe the baseline approaches in detail below:

\subsubsection{Zero-shot NER}
\label{section:zeroShot}
To perform Zero-shot transfer, we consider the mBERT model fine-tuned for NER task in English. We use the publicly available fine-tuned NER model\footnote{\label{nerExisting}\href{https://huggingface.co/Davlan/bert-base-multilingual-cased-ner-hrl}{Davlan/bert-base-multilingual-cased-ner-hrl}}\addtocounter{footnote}{-1}\addtocounter{Hfootnote}{-1} which is trained for NER in 10 high-resource languages (English, Arabic, Chinese, and some European languages). We directly test the performance of this model on {\dsname} large-test dataset (Bn, Gu, Hi, Kn, Ml, Mr, Pa, Ta, Te) and {\dsname} small-test datasets (As, Or) respectively.

\begin{table}[hbt!]
\centering
\small
\begin{tabular}{lcrrr} 
\toprule
\multirow{2}{*}{\textbf{Langs.}} & \multirow{2}{*}{\textbf{ZS}} & 
\multicolumn{1}{c}{\textbf{Teacher}} & \multicolumn{2}{c}{\textbf{Mined Data}} \\ 
\cmidrule{4-5}
 & & \multicolumn{1}{c}{\textbf{Student}} & \multicolumn{1}{c}{\textbf{GIZA++}} & 
 \multicolumn{1}{c}{\textbf{\begin{tabular}[c]{@{}c@{}}Awesome\\Align\end{tabular}}} \\ 
\midrule
bn & 64.83 & 63.07 & 79.35 & 81.02 \\
gu & 61.31  & 39.98 & 76.00 & 80.59 \\
hi & 71.77  & 73.62 & 80.44 & 82.69 \\
kn & 65.37  & 43.96 & 74.01 & 80.33 \\
ml & 70.47  & 70.97 & 72.35 & 81.49 \\
mr & 71.94  & 61.09 & 74.34 & 81.37 \\ 
pa & 58.03  & 44.90 & 70.91 & 71.51 \\
ta & 61.21  & 42.72 & 72.50 & 73.36 \\
te & 66.55  & 48.33 & 78.26 & 82.49 \\
\midrule
as & 25.40  & 13.03 & 13.04 & 45.37 \\
or & 1.71  & 0.22 & 18.77 & 25.01 \\ 
\midrule
\textbf{Mean} & 56.96  & 46.36 & 64.54 & \textbf{71.38} \\
\bottomrule
\end{tabular}
\caption{ F1 scores of various models on {\dsname} Test Set. Zero-Shot (ZS) and Teacher-Student models are described in Section \ref{section:zeroShot} and \ref{section:teacherStudent} respectively.}
\label{table:ai4bharatTestSetResults}
\end{table}

\subsubsection{Teacher-Student Learning}
\label{section:teacherStudent}
We use the publicly available fine-tuned NER model\footnotemark \: to create synthetic named entity annotated training data for the Indic languages. We annotate the Indic language portion of the Samanantar corpus using the above NER model. This synthetic labeled data is used to fine-tune for NER task in each of the languages respectively. 

\citet{wu-etal-2020-single} trained the student model to mimic the probability distribution of the entity labels by the teacher model. In our approach, we follow the \textit{Hard Labels} approach where we round the probability distribution of the entity labels into a one-hot labeling vector to guide the learning of the student model.

\subsection{Implementation Details}
We use the Huggingface library\footnote{\url{https://github.com/huggingface/transformers/tree/main/examples/pytorch/token-classification}} \cite{wolf-etal-2020-transformers}  to train our NER models. We use NVIDIA A100 Tensor Core GPU to run all the experiments. We use \texttt{bert-base-multilingual-cased} (169.05M) as the base pre-trained model in all our experiments. We tune hyper-parameters based on F1-Score on the validation set. We use the following range of values for selecting the best hyper-parameter. 
\begin{itemize}
\item Batch Size: 8, 16, 32
\item Learning Rate: 1e-3, 1e-4, 1e-5, 1e-6, 3e-3, 3e-4, 3e-5, 3e-6, 5e-3, 5e-4, 5e-5, 5e-6 
\end{itemize}
Once we obtain the best hyper-parameter, we fine-tune the model for $2$ epochs with $5$ different random seeds. We report the mean and standard deviation of the $5$ runs.

%% file: sections/results.tex
\begin{table}[hbt!]
\centering
\small
\begin{tabular}{lrr}
\toprule
\multicolumn{1}{c}{\textbf{Languages}} & \multicolumn{1}{c}{\textbf{Monolingual}} & \multicolumn{1}{c}{\textbf{Multilingual}} \\
\midrule
bn & 81.02 $\pm$ 0.40 & 80.74 $\pm$ 0.43 \\
gu & 80.59 $\pm$ 0.57 & 81.10 $\pm$ 0.39 \\
hi & 82.69 $\pm$ 0.45 & 82.93 $\pm$ 0.47 \\
kn & 80.33 $\pm$ 0.60 & 81.07 $\pm$ 0.55 \\
ml & 81.49 $\pm$ 0.15 & 81.13 $\pm$ 0.43 \\
mr & 81.37 $\pm$ 0.29 & 81.13 $\pm$ 0.47 \\
pa & 71.51 $\pm$ 0.59 & 71.81 $\pm$ 0.46 \\
ta & 73.36 $\pm$ 0.56 & 74.11 $\pm$ 0.46 \\
te & 82.49 $\pm$ 0.60 & 82.20 $\pm$ 0.31 \\
\midrule
as & 45.37 $\pm$ 2.66 & 60.19 $\pm$ 4.80 \\
or & 25.01 $\pm$ 1.22 & 25.91 $\pm$ 0.40 \\
\midrule
\textbf{Average} & 71.38 $\pm$ 0.69 & 72.94 $\pm$ 0.40 \\
\bottomrule
\end{tabular}
\caption{Comparison of Monolingual \textit{vs.} Multilingual Fine-tuning (F1 score). We report mean and standard deviation from 5 runs}
\label{tab:monoMulti}
\end{table}

\section{Results}
We now present the results from our experiments.

\begin{table}[!htb]
\centering
\small
\begin{tabular}{@{}lcccc@{}}
\toprule
\textbf{} & \multicolumn{1}{c}{\textbf{\texttt{PER}}} & \multicolumn{1}{c}{\textbf{\texttt{LOC}}} & \multicolumn{1}{c}{\textbf{\texttt{ORG}}} & \multicolumn{1}{c}{\textbf{Overall}}  \\ 
\midrule
bn & 77.63 & 84.29 & 73.25 & 80.06 \\
gu & 81.14 & 88.65 & 67.63 & 80.83 \\
hi & 82.31 & 89.37 & 74.03 & 83.27 \\
kn & 78.16 & 87.29 & 73.12 & 81.28 \\
ml & 84.49 & 87.85 & 61.49 & 81.67 \\
mr & 83.70 & 88.66 & 66.33 & 81.88 \\
pa & 76.26 & 77.95 & 55.68 & 72.08 \\
ta & 76.01 & 83.09 & 58.73 & 74.48 \\
te & 84.38 & 84.77 & 70.92 & 81.90 \\
\midrule
as & 75.00 & 54.55 & 57.14 & 62.50 \\
or & 41.78 & 21.40 & 13.39 & 26.42 \\
\bottomrule
\end{tabular}%
\caption{Entity-wise F1 score from the best multilingual model on {\dsname} Test Set. }
\label{table:ai4bharatTestSetEntityResults}
\end{table}

\begin{table*}[htb!]
\centering
\small
\begin{tabular}{@{}crrrrrr@{}}
\toprule
\textbf{Language} & \multicolumn{1}{c}{\textbf{\dsname}} & \multicolumn{1}{c}{\textbf{FIRE-2014}} & \multicolumn{1}{c}{\textbf{WikiANN}} & \multicolumn{1}{c}{\textbf{MultiCoNER}} & \multicolumn{1}{c}{\textbf{CFILT}} & \multicolumn{1}{c}{\textbf{MahaNER}} \\ \midrule
bn & 81.02 $\pm$ 0.40 & 35.68 $\pm$ 3.96 & 51.67 $\pm$ 1.24 & 26.12 $\pm$ 1.96 & - & - \\ 
gu & 80.59 $\pm$ 0.57 & - & 0.11 $\pm$ 0.12 & - & - & - \\ 
hi & 82.69 $\pm$ 0.45 & 47.23 $\pm$ 0.92 & 59.84 $\pm$ 1.25 & 41.85 $\pm$ 2.34 & 75.71 $\pm$ 0.67 & - \\ 
kn & 80.33 $\pm$ 0.60 & - & 2.73 $\pm$ 1.47 & - & - & - \\ 
ml & 81.49 $\pm$ 0.15 & 58.51 $\pm$ 1.13 & 62.59 $\pm$ 0.32 & - & - & - \\ 
mr & 81.37 $\pm$ 0.29 & - & 62.37 $\pm$ 1.12 & - & 58.41 $\pm$ 0.62 & 71.45 $\pm$ 1.44 \\ 
pa & 71.51 $\pm$ 0.59 & - & 0.7 $\pm$ 0.37 & - & - & - \\ 
ta & 73.36 $\pm$ 0.56 & 44.89 $\pm$ 0.94 & 49.15 $\pm$ 1.17 & - & - & - \\
te & 82.49 $\pm$ 0.60 & - & 49.28 $\pm$ 2.17 & - & - & - \\
\bottomrule
\end{tabular}
\caption{Comparison of models trained on different datasets and evaluated on \dsname-test set (F1 score).}
\label{table:big_human_all_dataset}
\end{table*}

\subsection{RQ1}
We now answer the question if \textit{the models trained using data from projection approach are better than cross-lingual zero-shot and teacher-student models?}

Table \ref{table:ai4bharatTestSetResults} reports the results from our experiments. Apart from Hindi, Malayalam, and, Marathi we observe relatively poor results for other Indic languages in the Zero-Shot setting. Zero-shot techniques perform quite well in high-resource languages like Hindi, scoring a respectable 75.96\%. However, for Assamese and Oriya languages the results are very poor. The Teacher-Student approach in comparison with the zero-shot approach gives very poor results.  

We observe that the models trained using the {\dsname-train} dataset give the best F1 scores across languages. In general, we observe better performance from data obtained using Awesome-align \cite{dou-neubig-2021-word} compared to GIZA++ \cite{och-ney-2003-systematic}. Moving forward, we choose the data obtained using Awesome-align \cite{dou-neubig-2021-word} in all our experiments.

\subsection*{\modelname: Multilingual Fine-tuning}
Multilingual fine-tuning on a downstream task has been shown to outperform language-specific fine-tuning in low-resource scenarios \cite{dhamecha-etal-2021-role}. We also fine-tune a multilingual model on the combined data of all languages in \dsname-train. We refer to this model as \modelname. Table \ref{tab:monoMulti} reports the results from our experiments. We observe that the multilingual model on average performs better than the monolingual models. 

It can also be seen that for extremely low-resource languages like Assamese, the multilingual model performs a lot better than the others with a jump in F1 score from $45.37$ to $60.19$.

\subsection{RQ2}
In this section, we answer the question \textit{if the models trained on the {\dsname-train} data fare better against models trained on other publicly available labeled datasets when tested on {\dsname-test} set?}

Table \ref{table:big_human_all_dataset} reports the results of our experiments. We observe that the model fine-tuned on {\dsname-train} data outperforms all other models by a significant margin indicating the utility of our labeled data. Only the models trained using CFILT-HiNER \cite{https://doi.org/10.48550/arxiv.2204.13743} and MahaNER \cite{litake-EtAl:2022:WILDRE6} obtain reasonable F1 on Hindi and Marathi. This underlines the importance of large, high-quality data and shows that projection methods can help to create such data at scale. 

\subsection{Error Analysis}
We observe that boundary error is our model's most common error type. The model sometimes identifies named entities partially. For example, in the case of Organization entities, our model only tags {\em A B C} as an organization entity when the correct entity phrase is, say, {\em A B C Limited}. Similarly, for Location entities, our model only tags {\em A B} as location entity when the correct entity phrase is {\em A B Hospital}. This could be attributed to some of the boundary errors present in our training data due to alignment errors.

%% file: sections/conclusion.tex
\section{Conclusion}
We take a major step towards creating publicly available, open datasets and open-source models for named entity recognition in Indic languages. We introduce {\dsname}, the largest entity recognition corpora for 11 Indic languages containing more than $100K$  training sentences per language, and covering 11 of the 22 languages listed in the Indian constitution. \dsname also includes manually labelled test set for $9$ Indic languages. We also build IndicNER, an mBERT based multilingual named entity recognition model for  11 Indic languages. We also provide baseline results on our test set along with a qualitative analysis of the model performance. The datasets and models will be available publicly under open-source licenses. We hope the dataset will spur innovations in entity recognition and its downstream applications in the Indian NLP space. 

\section*{Acknowledgements}
We would like to thank the Ministry of Electronics and Information Technology\footnote{\url{https://www.meity.gov.in/}} of the Government of India for their generous grant through the Digital India Bhashini project\footnote{\url{https://www.bhashini.gov.in/}}. We also thank the Centre for Development of Advanced Computing\footnote{\url{ https://www.cdac.in/index.aspx?id=pune}} for providing compute time on the Param Siddhi Supercomputer. We also thank Nilekani Philanthropies for their generous grant towards building datasets, models, tools and resources for Indic languages. We also thank Microsoft for their grant to support research on Indic languages.  Most importantly, we would like to thank Archana Mhaske, Anil Mhaske, Sangeeta Rajagopal, Vindhya DS, Nitin Kedia, Yash Madhani, Kabir Ahuja, Shallu Rani, Armin Virk and Gowtham Ramesh for volunteering to annotate the testset.

%% file: sections/appendix.tex
\appendix

\label{sec:appendix}

\section{Issues with WikiAnn}
\label{sec:wikiann}

\begin{figure*}[!htb]
 \centering
 \includegraphics[width=0.9\textwidth]{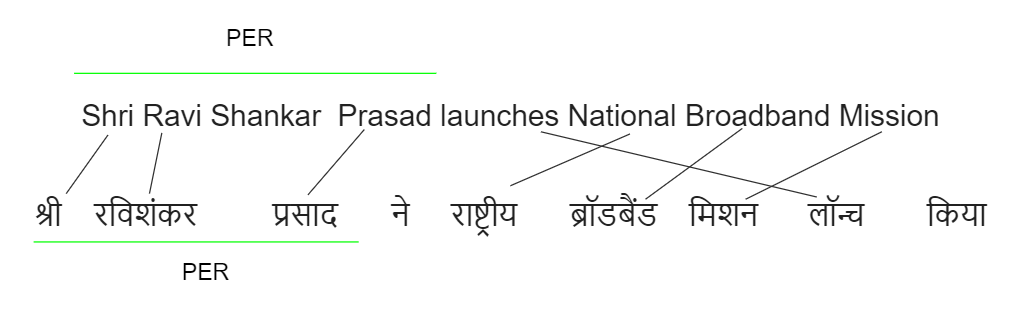}
 \caption{Correct Projection and Alignment using Awesome Align}
 \label{align_awesome_correct}
 \centering
 \includegraphics[width=0.9\textwidth]{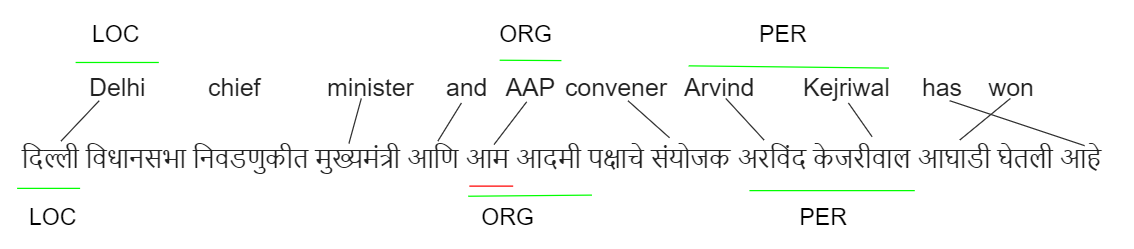}
 \caption{Incorrect Projection and Alignment using Awesome Align}
 \label{align_awesome_incorrect}
\end{figure*}

On manual inspection, the sentences in the Wiki dataset had a lot of issues. The ``sentences" were mostly just phrases and titles where more often than not, the entire thing would be considered a named entity. Such a skewed dataset can heavily influence the quality of a model trained on it. A few examples depicting the above issues are shown below.
\begin{itemize}
    \item \indicWords{120}
    \\B-LOC I-LOC I-LOC
    \item \indicWords{121}
    \\B-ORG I-ORG I-ORG I-ORG I-ORG
    \item \indicWords{122}
    \\B-ORG I-ORG I-ORG I-ORG
    \item \indicWords{123}
    \\B-LOC I-LOC I-LOC I-LOC
\end{itemize}

\section{Comparison English NER taggers}
\label{sec:eng_ner}

We compared many English NER taggers. The results are shown in Table \ref{tab:en-baselines}. 

\begin{table}[H]
\centering
\resizebox{0.45\textwidth}{!}{%
\begin{tabular}{llr}
\toprule
\multicolumn{1}{c}{\textbf{Model}} & \multicolumn{1}{c}{\textbf{Reference}} & \multicolumn{1}{c}{\textbf{F1}} \\
\midrule
Spacy NER     & \citet{schmitt2019replicable} & 91.60 \\
LUKE          & \citet{yamada2020luke} & 94.30 \\
BERT-base-NER & \citet{devlin-etal-2019-bert} & 91.30 \\
\bottomrule
\end{tabular}%
}
\caption{F1 scores for various off-the-shelf English models on CoNLL-2003 testset}
\label{tab:en-baselines}
\end{table}

\section{Examples of alignments generated by Awesome-align}
\label{sec:alignment_examples}

We now present a few examples from our projection method. Figure \ref{align_awesome_correct} presents examples of correct alignments and hence correct projections of NER tags. As can be seen, the alignment is fairly sparse and the model aligns only those words in which it is extremely confident. In this sentence, both words ``Ravi" and ``Shankar" had to be aligned to ``Ravishankar" in Hindi, but only ``Ravi" was aligned. But due to our range projection, the entire entity ``Shri Ravi Shankar Prasad" was projected successfully with the tag PERSON.

Figure \ref{align_awesome_incorrect} shows an example of incorrect word alignment using the awesome align method for word alignment. In this sentence, ``AAP" which is the abbreviated name of a political party is mapped only to ``Aam" in Marathi instead of the entire phrase ``Aam Aadmi pakshanche". This causes the projected entity to be only partially tagged with the entity type Organization.

\section{Comparison with Other Pre-Trained Language Models}

\begin{table}[!htb]
\centering
\small
\begin{tabular}{@{}lrrr@{}}
\toprule
\multicolumn{1}{c}{\textbf{Languages}} & \multicolumn{1}{c}{\textbf{mBERT}} & \multicolumn{1}{c}{\textbf{MuRIL}} & \multicolumn{1}{c}{\textbf{IndicBERT}}  \\ 
\midrule
bn & 80.74 $\pm$ 0.43 & 80.97 $\pm$ 0.28 & 81.02 $\pm$ 0.53 \\
gu & 81.10 $\pm$ 0.39 & 80.08 $\pm$ 0.27 & 80.34 $\pm$ 0.20 \\
hi & 82.93 $\pm$ 0.47 & 82.25 $\pm$ 0.28 & 82.40 $\pm$ 0.11 \\
kn & 81.07 $\pm$ 0.55 & 80.38 $\pm$ 0.38 & 80.74 $\pm$ 0.43 \\
ml & 81.13 $\pm$ 0.43 & 80.53 $\pm$ 0.44 & 80.45 $\pm$ 0.44 \\
mr & 81.13 $\pm$ 0.47 & 80.16 $\pm$ 0.28 & 80.52 $\pm$ 0.29 \\
pa & 71.81 $\pm$ 0.46 & 72.01 $\pm$ 0.26 & 71.66 $\pm$ 0.25 \\
ta & 74.11 $\pm$ 0.46 & 74.90 $\pm$ 3.87 & 74.85 $\pm$ 2.74 \\ 
te & 82.20 $\pm$ 0.31 & 81.83 $\pm$ 0.29 & 82.33 $\pm$ 0.50 \\ 
\midrule
as & 60.19 $\pm$ 4.80 & 66.03 $\pm$ 3.30 & 66.65 $\pm$ 3.73 \\
or & 25.91 $\pm$ 0.40 & 39.29 $\pm$ 0.60 & 39.44 $\pm$ 0.65 \\ 
\midrule
Average & 72.94 $\pm$ 0.40 & 74.40 $\pm$ 0.34 & 74.58 $\pm$ 0.55 \\
\bottomrule
\end{tabular}
\caption{Comparison of various pre-trained models fine-tuned on \dsname-train in a multilingual fashion.}
\label{tab:modelComparison}
\end{table}

Table \ref{tab:modelComparison} reports the performance of various pre-trained models fine-tuned on \dsname-train set in a multilingual fashion. We observe both MuRIL \cite{khanuja2021muril} and IndicBERT \cite{https://doi.org/10.48550/arxiv.2212.05409} outperform mBERT model.


\begin{table*}[!htb]
\centering
\small
\begin{tabular}{lrrrrrrrrrrr}
\toprule
 & \textbf{as} & \textbf{bn} & \textbf{gu} & \textbf{hi} & \textbf{kn} & \textbf{ml} & \textbf{mr} & \textbf{or} & \textbf{pa} & \textbf{ta} & \textbf{te} \\ 
\midrule
\textbf{Un-filtered} & 141K & 8.6M & 3M & 10M & 4M & 5.9M & 3.6M & 998K & 2.9M & 5.2M & 4.9M \\ 
\textbf{Filtered} & 10K & 966K & 475K & 999K & 474K & 720K & 457K & 197K & 465K & 500K & 510K \\ \bottomrule
\end{tabular}%
\caption{Filtering Statistics (Number of Sentences)}
\label{filt_table}
\end{table*}

%% file: main.bbl
\begin{thebibliography}{48}
\expandafter\ifx\csname natexlab\endcsname\relax\def\natexlab#1{#1}\fi

\bibitem[{Agerri et~al.(2018)Agerri, Chung, Aldabe, Aranberri, Labaka, and
  Rigau}]{agerri2018building}
Rodrigo Agerri, Yiling Chung, Itziar Aldabe, Nora Aranberri, Gorka Labaka, and
  German Rigau. 2018.
\newblock Building named entity recognition taggers via parallel corpora.
\newblock In \emph{Proceedings of the Eleventh International Conference on
  Language Resources and Evaluation (LREC 2018)}.

\bibitem[{Bansal et~al.(2022)Bansal, Ghorbani, Garg, Zhang, Cherry, Neyshabur,
  and Firat}]{bansal2022data}
Yamini Bansal, Behrooz Ghorbani, Ankush Garg, Biao Zhang, Colin Cherry, Behnam
  Neyshabur, and Orhan Firat. 2022.
\newblock Data scaling laws in nmt: The effect of noise and architecture.
\newblock In \emph{International Conference on Machine Learning}, pages
  1466--1482. PMLR.

\bibitem[{Benikova et~al.(2014)Benikova, Biemann, and
  Reznicek}]{benikova-etal-2014-nosta}
Darina Benikova, Chris Biemann, and Marc Reznicek. 2014.
\newblock \href
  {http://www.lrec-conf.org/proceedings/lrec2014/pdf/276_Paper.pdf}
  {{N}o{S}ta-{D} named entity annotation for {G}erman: Guidelines and dataset}.
\newblock In \emph{Proceedings of the Ninth International Conference on
  Language Resources and Evaluation ({LREC}'14)}, pages 2524--2531, Reykjavik,
  Iceland. European Language Resources Association (ELRA).

\bibitem[{Cohen(1960)}]{cohen1960coefficient}
Jacob Cohen. 1960.
\newblock A coefficient of agreement for nominal scales.
\newblock \emph{Educational and psychological measurement}, 20(1):37--46.

\bibitem[{Conneau et~al.(2020)Conneau, Khandelwal, Goyal, Chaudhary, Wenzek,
  Guzm{\'a}n, Grave, Ott, Zettlemoyer, and
  Stoyanov}]{conneau-etal-2020-unsupervised}
Alexis Conneau, Kartikay Khandelwal, Naman Goyal, Vishrav Chaudhary, Guillaume
  Wenzek, Francisco Guzm{\'a}n, Edouard Grave, Myle Ott, Luke Zettlemoyer, and
  Veselin Stoyanov. 2020.
\newblock \href {https://doi.org/10.18653/v1/2020.acl-main.747} {Unsupervised
  cross-lingual representation learning at scale}.
\newblock In \emph{Proceedings of the 58th Annual Meeting of the Association
  for Computational Linguistics}, pages 8440--8451, Online. Association for
  Computational Linguistics.

\bibitem[{Deleger et~al.(2012)Deleger, Li, Lingren, Kaiser, Molnar,
  Stoutenborough, Kouril, Marsolo, and Solti}]{interAnnotatorCite}
Louise Deleger, Qi~Li, Todd Lingren, Megan Kaiser, Katalin Molnar, Laura
  Stoutenborough, Michal Kouril, Keith Marsolo, and Imre Solti. 2012.
\newblock Building gold standard corpora for medical natural language
  processing tasks.
\newblock \emph{AMIA ... Annual Symposium proceedings / AMIA Symposium. AMIA
  Symposium}, 2012:144--153.

\bibitem[{Devlin et~al.(2019)Devlin, Chang, Lee, and
  Toutanova}]{devlin-etal-2019-bert}
Jacob Devlin, Ming-Wei Chang, Kenton Lee, and Kristina Toutanova. 2019.
\newblock \href {https://doi.org/10.18653/v1/N19-1423} {{BERT}: Pre-training of
  deep bidirectional transformers for language understanding}.
\newblock In \emph{Proceedings of the 2019 Conference of the North {A}merican
  Chapter of the Association for Computational Linguistics: Human Language
  Technologies, Volume 1 (Long and Short Papers)}, pages 4171--4186,
  Minneapolis, Minnesota. Association for Computational Linguistics.

\bibitem[{Dhamecha et~al.(2021)Dhamecha, Murthy, Bharadwaj, Sankaranarayanan,
  and Bhattacharyya}]{dhamecha-etal-2021-role}
Tejas Dhamecha, Rudra Murthy, Samarth Bharadwaj, Karthik Sankaranarayanan, and
  Pushpak Bhattacharyya. 2021.
\newblock \href {https://doi.org/10.18653/v1/2021.emnlp-main.675} {Role of
  {L}anguage {R}elatedness in {M}ultilingual {F}ine-tuning of {L}anguage
  {M}odels: {A} {C}ase {S}tudy in {I}ndo-{A}ryan {L}anguages}.
\newblock In \emph{Proceedings of the 2021 Conference on Empirical Methods in
  Natural Language Processing}, pages 8584--8595, Online and Punta Cana,
  Dominican Republic. Association for Computational Linguistics.

\bibitem[{Doddapaneni et~al.(2022)Doddapaneni, Aralikatte, Ramesh, Goyal,
  Khapra, Kunchukuttan, and Kumar}]{https://doi.org/10.48550/arxiv.2212.05409}
Sumanth Doddapaneni, Rahul Aralikatte, Gowtham Ramesh, Shreya Goyal, Mitesh~M.
  Khapra, Anoop Kunchukuttan, and Pratyush Kumar. 2022.
\newblock \href {https://doi.org/10.48550/ARXIV.2212.05409} {Indicxtreme: A
  multi-task benchmark for evaluating indic languages}.

\bibitem[{Dou and Neubig(2021)}]{dou-neubig-2021-word}
Zi-Yi Dou and Graham Neubig. 2021.
\newblock \href {https://doi.org/10.18653/v1/2021.eacl-main.181} {Word
  alignment by fine-tuning embeddings on parallel corpora}.
\newblock In \emph{Proceedings of the 16th Conference of the European Chapter
  of the Association for Computational Linguistics: Main Volume}, pages
  2112--2128, Online. Association for Computational Linguistics.

\bibitem[{Hu et~al.(2020)Hu, Ruder, Siddhant, Neubig, Firat, and
  Johnson}]{pmlr-v119-hu20b}
Junjie Hu, Sebastian Ruder, Aditya Siddhant, Graham Neubig, Orhan Firat, and
  Melvin Johnson. 2020.
\newblock \href {https://proceedings.mlr.press/v119/hu20b.html} {{XTREME}: A
  massively multilingual multi-task benchmark for evaluating cross-lingual
  generalisation}.
\newblock In \emph{Proceedings of the 37th International Conference on Machine
  Learning}, volume 119 of \emph{Proceedings of Machine Learning Research},
  pages 4411--4421. PMLR.

\bibitem[{Jain et~al.(2019)Jain, Paranjape, and Lipton}]{jain-etal-2019-entity}
Alankar Jain, Bhargavi Paranjape, and Zachary~C. Lipton. 2019.
\newblock \href {https://doi.org/10.18653/v1/D19-1100} {Entity projection via
  machine translation for cross-lingual {NER}}.
\newblock In \emph{Proceedings of the 2019 Conference on Empirical Methods in
  Natural Language Processing and the 9th International Joint Conference on
  Natural Language Processing (EMNLP-IJCNLP)}, pages 1083--1092, Hong Kong,
  China. Association for Computational Linguistics.

\bibitem[{Kakwani et~al.(2020)Kakwani, Kunchukuttan, Golla, N.C.,
  Bhattacharyya, Khapra, and Kumar}]{kakwani2020indicnlpsuite}
Divyanshu Kakwani, Anoop Kunchukuttan, Satish Golla, Gokul N.C., Avik
  Bhattacharyya, Mitesh~M. Khapra, and Pratyush Kumar. 2020.
\newblock {IndicNLPSuite: Monolingual Corpora, Evaluation Benchmarks and
  Pre-trained Multilingual Language Models for Indian Languages}.
\newblock In \emph{Findings of EMNLP}.

\bibitem[{Karthikeyan et~al.(2020)Karthikeyan, Wang, Mayhew, and
  Roth}]{K2020Cross-Lingual}
K~Karthikeyan, Zihan Wang, Stephen Mayhew, and Dan Roth. 2020.
\newblock \href {https://openreview.net/forum?id=HJeT3yrtDr} {Cross-lingual
  ability of multilingual bert: An empirical study}.
\newblock In \emph{International Conference on Learning Representations}.

\bibitem[{Khanuja et~al.(2021)Khanuja, Bansal, Mehtani, Khosla, Dey, Gopalan,
  Margam, Aggarwal, Nagipogu, Dave, Gupta, Gali, Subramanian, and
  Talukdar}]{khanuja2021muril}
Simran Khanuja, Diksha Bansal, Sarvesh Mehtani, Savya Khosla, Atreyee Dey,
  Balaji Gopalan, Dilip~Kumar Margam, Pooja Aggarwal, Rajiv~Teja Nagipogu,
  Shachi Dave, Shruti Gupta, Subhash Chandra~Bose Gali, Vish Subramanian, and
  Partha Talukdar. 2021.
\newblock \href {http://arxiv.org/abs/2103.10730} {Muril: Multilingual
  representations for indian languages}.

\bibitem[{Kumar et~al.(2022)Kumar, Shrotriya, Sahu, Dabre, Puduppully,
  Kunchukuttan, Mishra, Khapra, and Kumar}]{kumar-etal-2022-indicnlg}
Aman Kumar, Himani Shrotriya, Prachi Sahu, Raj Dabre, Ratish Puduppully, Anoop
  Kunchukuttan, Amogh Mishra, Mitesh~M. Khapra, and Pratyush Kumar. 2022.
\newblock Indicnlg benchmark: Multilingual datasets for diverse nlg tasks in
  indic languages.
\newblock In \emph{EMNLP}.

\bibitem[{Lalitha~Devi et~al.(2014)Lalitha~Devi, RK~Rao, C.S, and
  Sundar~Ram}]{FIRE2014}
Sobha Lalitha~Devi, Pattabhi RK~Rao, Malarkodi C.S, and R~Vijay Sundar~Ram.
  2014.
\newblock {Indian Language NER Annotated FIRE 2014 Corpus (FIRE 2014 NER
  Corpus)}.
\newblock In \emph{In Named-Entity Recognition Indian Languages FIRE 2014
  Evaluation Track}.

\bibitem[{Li et~al.(2020)Li, Sun, Meng, Liang, Wu, and Li}]{li-etal-2020-dice}
Xiaoya Li, Xiaofei Sun, Yuxian Meng, Junjun Liang, Fei Wu, and Jiwei Li. 2020.
\newblock \href {https://doi.org/10.18653/v1/2020.acl-main.45} {Dice loss for
  data-imbalanced {NLP} tasks}.
\newblock In \emph{Proceedings of the 58th Annual Meeting of the Association
  for Computational Linguistics}, pages 465--476, Online. Association for
  Computational Linguistics.

\bibitem[{Litake et~al.(2022)Litake, Sabane, Patil, Ranade, and
  Joshi}]{litake-EtAl:2022:WILDRE6}
Onkar Litake, Maithili~Ravindra Sabane, Parth~Sachin Patil, Aparna~Abhijeet
  Ranade, and Raviraj Joshi. 2022.
\newblock L3cube-mahaner: A marathi named entity recognition dataset and bert
  models.
\newblock In \emph{Proceedings of The WILDRE-6 Workshop within the 13th
  Language Resources and Evaluation Conference}, pages 29--34, Marseille,
  France. European Language Resources Association.

\bibitem[{Madhani et~al.(2022)Madhani, Parthan, Bedekar, Khapra, Seshadri,
  Kunchukuttan, Kumar, and Khapra}]{https://doi.org/10.48550/arxiv.2205.03018}
Yash Madhani, Sushane Parthan, Priyanka Bedekar, Ruchi Khapra, Vivek Seshadri,
  Anoop Kunchukuttan, Pratyush Kumar, and Mitesh~M. Khapra. 2022.
\newblock \href {https://doi.org/10.48550/ARXIV.2205.03018} {Aksharantar:
  Towards building open transliteration tools for the next billion users}.

\bibitem[{Malmasi et~al.(2022)Malmasi, Fetahu, Fang, Kar, and
  Rokhlenko}]{shervin-eta-al-2022-semeval}
Shervin Malmasi, Besnik Fetahu, Anjie Fang, Sudipta Kar, and Oleg Rokhlenko.
  2022.
\newblock {S}em{E}val-2022 task 11: Multiconer multilingual complex named
  entity recognition.
\newblock In \emph{Proceedings of the 16th International Workshop on Semantic
  Evaluation (SemEval-2022)}, Online. Association for Computational
  Linguistics.

\bibitem[{Mayhew et~al.(2017)Mayhew, Tsai, and Roth}]{mayhew2017cheap}
Stephen Mayhew, Chen-Tse Tsai, and Dan Roth. 2017.
\newblock Cheap translation for cross-lingual named entity recognition.
\newblock In \emph{Proceedings of the 2017 conference on empirical methods in
  natural language processing}, pages 2536--2545.

\bibitem[{Murthy et~al.(2022)Murthy, Bhattacharjee, Sharnagat, Khatri, Kanojia,
  and Bhattacharyya}]{https://doi.org/10.48550/arxiv.2204.13743}
Rudra Murthy, Pallab Bhattacharjee, Rahul Sharnagat, Jyotsana Khatri, Diptesh
  Kanojia, and Pushpak Bhattacharyya. 2022.
\newblock \href {https://doi.org/10.48550/ARXIV.2204.13743} {Hiner: A large
  hindi named entity recognition dataset}.

\bibitem[{Murthy et~al.(2018)Murthy, Kunchukuttan, and
  Bhattacharyya}]{murthy-etal-2018-judicious}
Rudra Murthy, Anoop Kunchukuttan, and Pushpak Bhattacharyya. 2018.
\newblock \href {https://doi.org/10.18653/v1/P18-2064} {Judicious selection of
  training data in assisting language for multilingual neural {NER}}.
\newblock In \emph{Proceedings of the 56th Annual Meeting of the Association
  for Computational Linguistics (Volume 2: Short Papers)}, pages 401--406,
  Melbourne, Australia. Association for Computational Linguistics.

\bibitem[{Och and Ney(2003)}]{och-ney-2003-systematic}
Franz~Josef Och and Hermann Ney. 2003.
\newblock \href {https://doi.org/10.1162/089120103321337421} {A systematic
  comparison of various statistical alignment models}.
\newblock \emph{Computational Linguistics}, 29(1):19--51.

\bibitem[{Pan et~al.(2017)Pan, Zhang, May, Nothman, Knight, and
  Ji}]{pan2017cross}
Xiaoman Pan, Boliang Zhang, Jonathan May, Joel Nothman, Kevin Knight, and Heng
  Ji. 2017.
\newblock Cross-lingual name tagging and linking for 282 languages.
\newblock In \emph{Proceedings of the 55th Annual Meeting of the Association
  for Computational Linguistics (Volume 1: Long Papers)}, pages 1946--1958.

\bibitem[{Pathak et~al.(2022)Pathak, Nandi, and
  Sarmah}]{pathak-etal-2022-asner}
Dhrubajyoti Pathak, Sukumar Nandi, and Priyankoo Sarmah. 2022.
\newblock \href {https://aclanthology.org/2022.lrec-1.706} {{A}s{NER} -
  annotated dataset and baseline for {A}ssamese named entity recognition}.
\newblock In \emph{Proceedings of the Thirteenth Language Resources and
  Evaluation Conference}, pages 6571--6577, Marseille, France. European
  Language Resources Association.

\bibitem[{Pires et~al.(2019)Pires, Schlinger, and
  Garrette}]{pires-etal-2019-multilingual}
Telmo Pires, Eva Schlinger, and Dan Garrette. 2019.
\newblock \href {https://doi.org/10.18653/v1/P19-1493} {How multilingual is
  multilingual {BERT}?}
\newblock In \emph{Proceedings of the 57th Annual Meeting of the Association
  for Computational Linguistics}, pages 4996--5001, Florence, Italy.
  Association for Computational Linguistics.

\bibitem[{Pradhan et~al.(2013)Pradhan, Moschitti, Xue, Ng, Bj{\"o}rkelund,
  Uryupina, Zhang, and Zhong}]{pradhan-etal-2013-towards}
Sameer Pradhan, Alessandro Moschitti, Nianwen Xue, Hwee~Tou Ng, Anders
  Bj{\"o}rkelund, Olga Uryupina, Yuchen Zhang, and Zhi Zhong. 2013.
\newblock \href {https://aclanthology.org/W13-3516} {Towards robust linguistic
  analysis using {O}nto{N}otes}.
\newblock In \emph{Proceedings of the Seventeenth Conference on Computational
  Natural Language Learning}, pages 143--152, Sofia, Bulgaria. Association for
  Computational Linguistics.

\bibitem[{Rahimi et~al.(2019)Rahimi, Li, and Cohn}]{rahimi-etal-2019-massively}
Afshin Rahimi, Yuan Li, and Trevor Cohn. 2019.
\newblock \href {https://doi.org/10.18653/v1/P19-1015} {Massively multilingual
  transfer for {NER}}.
\newblock In \emph{Proceedings of the 57th Annual Meeting of the Association
  for Computational Linguistics}, pages 151--164, Florence, Italy. Association
  for Computational Linguistics.

\bibitem[{Ramesh et~al.(2022)Ramesh, Doddapaneni, Bheemaraj, Jobanputra, AK,
  Sharma, Sahoo, Diddee, J, Kakwani, Kumar, Pradeep, Nagaraj, Deepak, Raghavan,
  Kunchukuttan, Kumar, and Khapra}]{ramesh-etal-2022-samanantar}
Gowtham Ramesh, Sumanth Doddapaneni, Aravinth Bheemaraj, Mayank Jobanputra,
  Raghavan AK, Ajitesh Sharma, Sujit Sahoo, Harshita Diddee, Mahalakshmi J,
  Divyanshu Kakwani, Navneet Kumar, Aswin Pradeep, Srihari Nagaraj, Kumar
  Deepak, Vivek Raghavan, Anoop Kunchukuttan, Pratyush Kumar, and
  Mitesh~Shantadevi Khapra. 2022.
\newblock \href {https://doi.org/10.1162/tacl_a_00452} {Samanantar: The largest
  publicly available parallel corpora collection for 11 indic languages}.
\newblock \emph{Transactions of the Association for Computational Linguistics},
  10:145--162.

\bibitem[{Ringland et~al.(2019)Ringland, Dai, Hachey, Karimi, Paris, and
  Curran}]{ringland-etal-2019-nne}
Nicky Ringland, Xiang Dai, Ben Hachey, Sarvnaz Karimi, Cecile Paris, and
  James~R. Curran. 2019.
\newblock \href {https://doi.org/10.18653/v1/P19-1510} {{NNE}: A dataset for
  nested named entity recognition in {E}nglish newswire}.
\newblock In \emph{Proceedings of the 57th Annual Meeting of the Association
  for Computational Linguistics}, pages 5176--5181, Florence, Italy.
  Association for Computational Linguistics.

\bibitem[{Ruder et~al.(2021)Ruder, Constant, Botha, Siddhant, Firat, Fu, Liu,
  Hu, Garrette, Neubig, and Johnson}]{ruder-etal-2021-xtreme}
Sebastian Ruder, Noah Constant, Jan Botha, Aditya Siddhant, Orhan Firat, Jinlan
  Fu, Pengfei Liu, Junjie Hu, Dan Garrette, Graham Neubig, and Melvin Johnson.
  2021.
\newblock \href {https://doi.org/10.18653/v1/2021.emnlp-main.802}
  {{XTREME}-{R}: Towards more challenging and nuanced multilingual evaluation}.
\newblock In \emph{Proceedings of the 2021 Conference on Empirical Methods in
  Natural Language Processing}, pages 10215--10245, Online and Punta Cana,
  Dominican Republic. Association for Computational Linguistics.

\bibitem[{Schmitt et~al.(2019)Schmitt, Kubler, Robert, Papadakis, and
  LeTraon}]{schmitt2019replicable}
Xavier Schmitt, Sylvain Kubler, J{\'e}r{\'e}my Robert, Mike Papadakis, and Yves
  LeTraon. 2019.
\newblock A replicable comparison study of ner software: Stanfordnlp, nltk,
  opennlp, spacy, gate.
\newblock In \emph{2019 Sixth International Conference on Social Networks
  Analysis, Management and Security (SNAMS)}, pages 338--343. IEEE.

\bibitem[{Shah et~al.(2010)Shah, Lin, Gershman, and
  Frederking}]{shah2010synergy}
Rushin Shah, Bo~Lin, Anatole Gershman, and Robert Frederking. 2010.
\newblock Synergy: a named entity recognition system for resource-scarce
  languages such as swahili using online machine translation.
\newblock In \emph{Proceedings of the Second Workshop on African Language
  Technology (AfLaT 2010)}, pages 21--26.

\bibitem[{Tjong Kim~Sang(2002)}]{tjong-kim-sang-2002-introduction}
Erik~F. Tjong Kim~Sang. 2002.
\newblock \href {https://aclanthology.org/W02-2024} {Introduction to the
  {C}o{NLL}-2002 shared task: Language-independent named entity recognition}.
\newblock In \emph{{COLING}-02: The 6th Conference on Natural Language Learning
  2002 ({C}o{NLL}-2002)}.

\bibitem[{Tjong Kim~Sang and
  De~Meulder(2003{\natexlab{a}})}]{tsang2003introduction}
Erik~F. Tjong Kim~Sang and Fien De~Meulder. 2003{\natexlab{a}}.
\newblock \href {https://www.aclweb.org/anthology/W03-0419} {Introduction to
  the {C}o{NLL}-2003 shared task: Language-independent named entity
  recognition}.
\newblock In \emph{Proceedings of the Seventh Conference on Natural Language
  Learning at {HLT}-{NAACL} 2003}, pages 142--147.

\bibitem[{Tjong Kim~Sang and
  De~Meulder(2003{\natexlab{b}})}]{tjong-kim-sang-de-meulder-2003-introduction}
Erik~F. Tjong Kim~Sang and Fien De~Meulder. 2003{\natexlab{b}}.
\newblock \href {https://aclanthology.org/W03-0419} {Introduction to the
  {C}o{NLL}-2003 shared task: Language-independent named entity recognition}.
\newblock In \emph{Proceedings of the Seventh Conference on Natural Language
  Learning at {HLT}-{NAACL} 2003}, pages 142--147.

\bibitem[{Wang et~al.(2021)Wang, Jiang, Bach, Wang, Huang, Huang, and
  Tu}]{wang-etal-2021-automated}
Xinyu Wang, Yong Jiang, Nguyen Bach, Tao Wang, Zhongqiang Huang, Fei Huang, and
  Kewei Tu. 2021.
\newblock \href {https://doi.org/10.18653/v1/2021.acl-long.206} {Automated
  concatenation of embeddings for structured prediction}.
\newblock In \emph{Proceedings of the 59th Annual Meeting of the Association
  for Computational Linguistics and the 11th International Joint Conference on
  Natural Language Processing (Volume 1: Long Papers)}, pages 2643--2660,
  Online. Association for Computational Linguistics.

\bibitem[{Wolf et~al.(2020)Wolf, Debut, Sanh, Chaumond, Delangue, Moi, Cistac,
  Rault, Louf, Funtowicz, Davison, Shleifer, von Platen, Ma, Jernite, Plu, Xu,
  Le~Scao, Gugger, Drame, Lhoest, and Rush}]{wolf-etal-2020-transformers}
Thomas Wolf, Lysandre Debut, Victor Sanh, Julien Chaumond, Clement Delangue,
  Anthony Moi, Pierric Cistac, Tim Rault, Remi Louf, Morgan Funtowicz, Joe
  Davison, Sam Shleifer, Patrick von Platen, Clara Ma, Yacine Jernite, Julien
  Plu, Canwen Xu, Teven Le~Scao, Sylvain Gugger, Mariama Drame, Quentin Lhoest,
  and Alexander Rush. 2020.
\newblock \href {https://doi.org/10.18653/v1/2020.emnlp-demos.6} {Transformers:
  State-of-the-art natural language processing}.
\newblock In \emph{Proceedings of the 2020 Conference on Empirical Methods in
  Natural Language Processing: System Demonstrations}, pages 38--45, Online.
  Association for Computational Linguistics.

\bibitem[{Wu et~al.(2020)Wu, Lin, Karlsson, Lou, and
  Huang}]{wu-etal-2020-single}
Qianhui Wu, Zijia Lin, B{\"o}rje Karlsson, Jian-Guang Lou, and Biqing Huang.
  2020.
\newblock \href {https://doi.org/10.18653/v1/2020.acl-main.581}
  {Single-/multi-source cross-lingual {NER} via teacher-student learning on
  unlabeled data in target language}.
\newblock In \emph{Proceedings of the 58th Annual Meeting of the Association
  for Computational Linguistics}, pages 6505--6514, Online. Association for
  Computational Linguistics.

\bibitem[{Wu and Dredze(2019)}]{wu-dredze-2019-beto}
Shijie Wu and Mark Dredze. 2019.
\newblock \href {https://doi.org/10.18653/v1/D19-1077} {Beto, bentz, becas: The
  surprising cross-lingual effectiveness of {BERT}}.
\newblock In \emph{Proceedings of the 2019 Conference on Empirical Methods in
  Natural Language Processing and the 9th International Joint Conference on
  Natural Language Processing (EMNLP-IJCNLP)}, pages 833--844, Hong Kong,
  China. Association for Computational Linguistics.

\bibitem[{Xie et~al.(2018)Xie, Yang, Neubig, Smith, and
  Carbonell}]{xie-etal-2018-neural}
Jiateng Xie, Zhilin Yang, Graham Neubig, Noah~A. Smith, and Jaime Carbonell.
  2018.
\newblock \href {https://doi.org/10.18653/v1/D18-1034} {Neural cross-lingual
  named entity recognition with minimal resources}.
\newblock In \emph{Proceedings of the 2018 Conference on Empirical Methods in
  Natural Language Processing}, pages 369--379, Brussels, Belgium. Association
  for Computational Linguistics.

\bibitem[{Yamada et~al.(2020)Yamada, Asai, Shindo, Takeda, and
  Matsumoto}]{yamada2020luke}
Ikuya Yamada, Akari Asai, Hiroyuki Shindo, Hideaki Takeda, and Yuji Matsumoto.
  2020.
\newblock Luke: deep contextualized entity representations with entity-aware
  self-attention.
\newblock \emph{arXiv preprint arXiv:2010.01057}.

\bibitem[{Yarowsky and Ngai(2001{\natexlab{a}})}]{yarowsky-ngai-2001-inducing}
David Yarowsky and Grace Ngai. 2001{\natexlab{a}}.
\newblock \href {https://aclanthology.org/N01-1026} {Inducing multilingual
  {POS} taggers and {NP} bracketers via robust projection across aligned
  corpora}.
\newblock In \emph{Second Meeting of the North {A}merican Chapter of the
  Association for Computational Linguistics}.

\bibitem[{Yarowsky and
  Ngai(2001{\natexlab{b}})}]{Yarowsky:2001:IMP:1073336.1073362}
David Yarowsky and Grace Ngai. 2001{\natexlab{b}}.
\newblock {Inducing Multilingual POS Taggers and NP Bracketers via Robust
  Projection Across Aligned Corpora}.
\newblock In \emph{Proceedings of the Second Meeting of the North American
  Chapter of the Association for Computational Linguistics on Language
  Technologies}.

\bibitem[{Yarowsky et~al.(2001)Yarowsky, Ngai, and
  Wicentowski}]{yarowsky-etal-2001-inducing}
David Yarowsky, Grace Ngai, and Richard Wicentowski. 2001.
\newblock \href {https://aclanthology.org/H01-1035} {Inducing multilingual text
  analysis tools via robust projection across aligned corpora}.
\newblock In \emph{Proceedings of the First International Conference on Human
  Language Technology Research}.

\bibitem[{Zhang et~al.(2016)Zhang, Zhang, Pan, Feng, Ji, and
  Xu}]{zhang-etal-2016-bitext}
Dongxu Zhang, Boliang Zhang, Xiaoman Pan, Xiaocheng Feng, Heng Ji, and Weiran
  Xu. 2016.
\newblock \href {https://aclanthology.org/C16-1045} {Bitext name tagging for
  cross-lingual entity annotation projection}.
\newblock In \emph{Proceedings of {COLING} 2016, the 26th International
  Conference on Computational Linguistics: Technical Papers}, pages 461--470,
  Osaka, Japan. The COLING 2016 Organizing Committee.

\end{thebibliography}
